\documentclass{article} 
\usepackage[final]{acl}
\usepackage{times}
\usepackage{amsmath}
\usepackage{hyperref}
\usepackage{url}
\usepackage{graphicx}
\usepackage{xcolor}
\usepackage{booktabs}
\usepackage{subcaption}
\usepackage{arydshln}
\title{Large Language Models are In-context Teachers \\for Knowledge Reasoning}

\author{
  \textbf{Jiachen Zhao\textsuperscript{1}},
  \textbf{Zonghai Yao\textsuperscript{2}},
  \textbf{Zhichao Yang\textsuperscript{2}},
  \textbf{Hong Yu\textsuperscript{2}},
\\
  \textsuperscript{1}Northeastern University,
  \textsuperscript{2}University of Massachusetts Amherst
\\
  \texttt{
   {zhao.jiach@northeastern.edu}
  }
}
\newcommand{\zjc}[1]{\textcolor{blue}{[J:#1]}}
\newcommand{\selfexp}{\textsf{Self-Explain}}
\newcommand{\teachback}{\textsf{Teach-Back}}
\begin{document}

\maketitle

\begin{abstract}\looseness=-1

In this work, we study \textit{in-context teaching} (ICT), where a teacher provides in-context example rationales to teach a student to reason over unseen cases. Human teachers are usually required to craft in-context demonstrations, which are costly and have high variance.  We ask whether a large language model (LLM) can serve as a more effective in-context teacher for itself or other LLMs, compared to humans. Inspired by the Encoding Specificity Hypothesis from human episodic memory, we hypothesize that in-context exemplars crafted by the teacher should match the training data of the student. This hypothesis motivates us to propose \selfexp{} where an LLM's self-elicited explanations are used as in-context demonstrations for prompting it as they are generalized from the model's training examples. \selfexp{} is shown to significantly outperform using human-crafted exemplars and other baselines. 

Furthermore, we reveal that for ICT,  rationales from different teacher LLMs or human experts that more resemble the student LLM's self-explanations are better in-context demonstrations. This supports our encoding specificity hypothesis. We then propose \teachback{} that aligns a teacher LLM with the student to enhance the ICT performance. For example, \teachback{} enables a 7B model to teach the much larger GPT-3.5 in context, surpassing human teachers by around 5\% in test accuracy on medical question answering.



\end{abstract}

\begin{figure*}[t]
    \centering
    \includegraphics[scale=0.47]{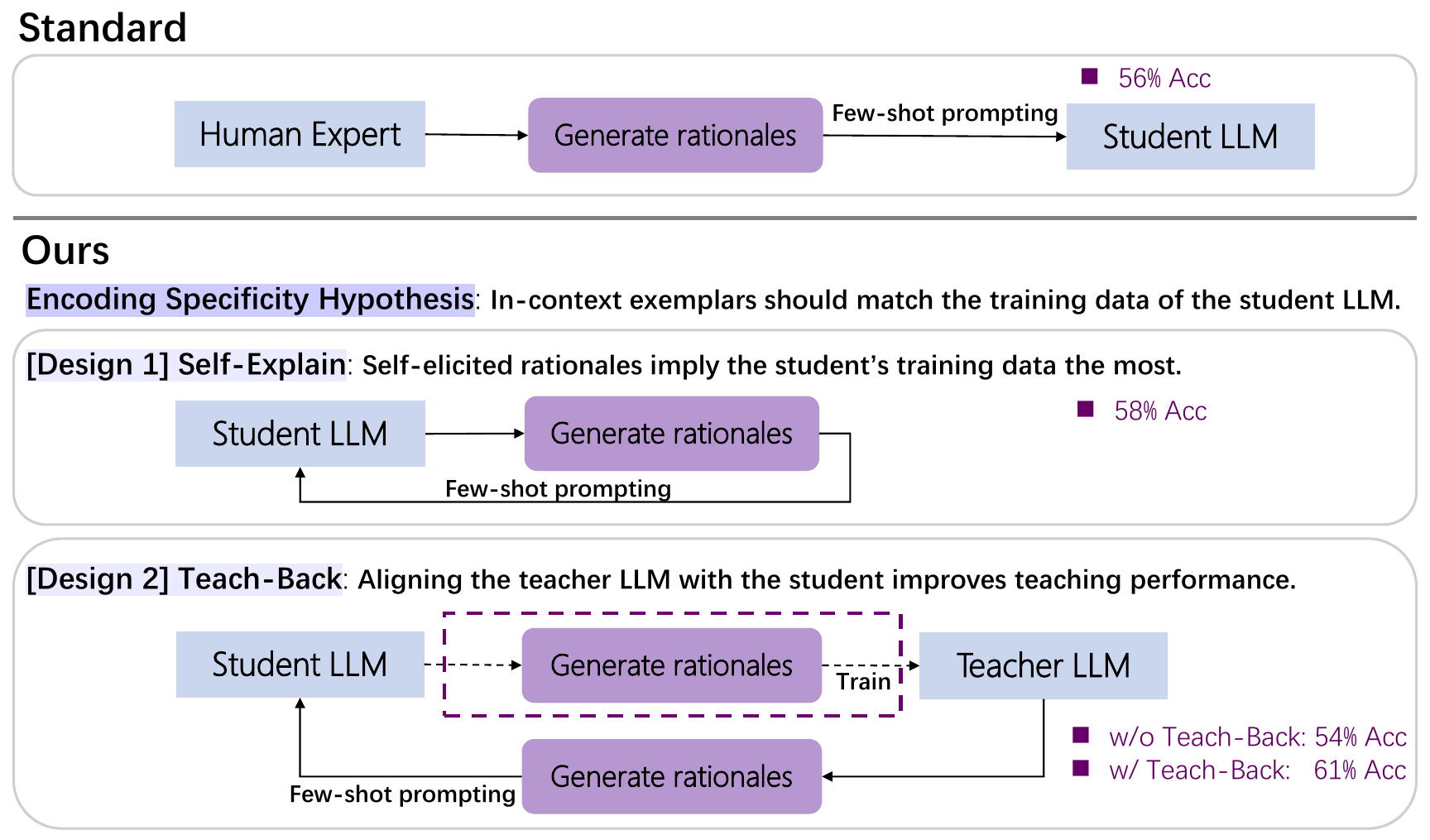}
    \caption{Overview of our approaches. Existing few-shot CoT prompting methods rely on human experts to craft rationales as in-context demonstrations.  We propose Encoding Specificity Hypothesis to make large language models better in-context teachers than humans.  We accordingly design \selfexp{} for an LLM itself to be the teacher and \teachback{} to improve the LLM's capability of teaching another model in context.
    }\vspace{-5mm}
    \label{fig:header}
\end{figure*}

\section{Introduction}
\looseness=-1
Knowledge reasoning, different from numerical reasoning, requires large language models (LLMs) to deduce the association between questions and answers that do not usually explicitly appear in the training corpus, although LLMs may have memorized all the facts involved in the question. 
Such a \textit{compositionality gap}~\citep{press-etal-2023-measuring} between testing and pretraining makes knowledge reasoning difficult and beyond mere fact retrieval. However, LLMs have demonstrated impressive knowledge reasoning performance on diverse tasks~\citep{DBLP:conf/nips/Wei0SBIXCLZ22} with few-shot prompting.  Exemplars of reasoning are provided in the prompt as context to teach LLMs to reason through in-context learning (ICL)~\citep{brown2020language} at inference. LLMs will generate intermediate reasoning steps (known as Chain-of-Thought (CoT)\footnote{We use ``CoT'' and ``rationale'' interchangeably to refer to reasoning paths.}~\citep{DBLP:conf/nips/Wei0SBIXCLZ22}) for deducing the test cases.







\looseness=-10000
Standard few-shot CoT prompting requires humans to first craft high-quality demonstrations of reasoning for LLMs, as depicted in the upper part of Figure~\ref{fig:header}. However, this may bring some issues. On the one hand, in professional domains such as medicine, experts like physicians are needed to produce fine-grained rationales with correct jargon, which is time-consuming and expensive~\citep{pal2022medmcqa,Yang2023.10.26.23297629}. On the other hand, different from labels, rationales can be phrased in varied ways, while all being correct~\citep{yao-etal-2023-human}. Collecting reasoning examples through crowd-sourcing can thus have great uncertainty~\citep{gebreegziabher2023patat}. The constructed rationales heavily depend on human annotators' own experience and thus, may be very subjective~\citep{lee2022evaluating}.


\looseness=-1000
More fundamentally, there is a limited understanding of the principles behind constructing effective rationale exemplars for in-context learning. 
Currently, the majority of works depend on human-crafted demonstrations (usually by professionals) that are based on some heuristic rules~\citep{DBLP:conf/iclr/FuPSCK23,zhou2022least,DBLP:conf/iclr/KhotTFF0CS23}. 
However, it is unclear whether those sophisticated rationales crafted by humans are equally the most sensible to LLMs.
Demonstrations of rationales from humans may not always be helpful~\citep{yao-etal-2023-human}, although they are often assumed to be gold standards~\citep{muller2021designing}.



\looseness=-1000
Therefore, we are motivated to ask, \textbf{can an LLM teach itself or other models through in-context learning for knowledge reasoning, preferably better than humans}? 
We consider a generic framework of \textit{in-context teaching} (ICT), where a teacher (e.g., human or LLM) constructs example rationales that are then used as in-context demonstrations to prompt a student LLM.  

In terms of how to construct exemplars for effective ICT, we consider the \textbf{Encoding Specificity Hypothesis}~\citep{tulving1972episodic}, which is initially proposed for retrieval from human episodic memory. The hypothesis postulates that the context during recalling information from episodic memory should match the context during encoding. As inspired by the convergence between memory and attention module in language models~\citep{ramsauer2020hopfield,bricken2021attention,zhao2023incontext}, for few-shot prompting, we similarly hypothesize that in-context exemplars at test time should match the encoded rationales from the student's training corpora related to the test domain. For example, when the student reasons over medical questions at inference, in-context rationales are expected to be phrased similarly to examples in the medical corpus learned by the student model during training.  

\looseness=-1000
The encoding specificity hypothesis can be easily satisfied when the teacher model is the student model itself. We directly prompt the student model to \textit{explain} the given answer to a question sampled from the same dataset as the test data as inspired by learning theory in cognitive science~\citep{chi1989self}. 
Those elicited self-explanations can represent the model's encoded knowledge for the test task and are then used as in-context demonstrations at inference. 
We refer to this approach as \selfexp{}. 
On the other hand, when the teacher model is different from the student model (e.g., using a weak and small model to teach a much larger model~\citep{burns2023weak}), we first let the teacher model learn from the student's self-explanations before eliciting the teacher's explanations (see the lower part of Figure~\ref{fig:header}). 
We refer to this method as \teachback{}, which is how healthcare providers (i.e., teachers) reduce the communication gap with patients (i.e., student) for effective health education~\citep{talevski2020teach}. 


Our experimental results provide sources of evidence for our encoding specificity hypothesis. We find that the student model itself tends to be the best in-context teacher for it, surpassing human teachers or other LLM teachers (w/o \teachback{}).  
Our experiments across models of different sizes and reasoning abilities suggest that for ICT, larger and stronger models are not necessarily better in-context teachers, though they may produce more reliable rationales.  We show that in-context reasoning examples that more resemble the student's self-explanations can lead to better student performance. This also supports our encoding specificity hypothesis.
Furthermore, applying \teachback{} can significantly improve the ICT capability of a teacher model for the student model and even outperform \selfexp{} depending on the teacher model. 
For example, \teachback{} enables a small deployable 7B model to teach the much larger GPT-3.5 in context, surpassing human teachers by around 5\% in test accuracy on medical question answering.

In summary, our contributions are mainly in three folds: 
\begin{itemize}
   \item We investigate in-context teaching for knowledge reasoning, where a teacher provides in-context demonstrations to teach a student model to reason. We propose the Encoding Specificity Hypothesis as the guideline for composing in-context exemplars and provide sufficient evidence for our hypothesis.
   \item We propose a new way of eliciting rationales from an LLM by prompting it to explain question-answer pairs. We then propose \selfexp{} prompting to use an LLM's self-explanations as its in-context exemplars, which outperforms using human-crafted CoTs. 
   \item Our experiments suggest that in-context exemplars of rationales from LLM teachers or human teachers that more resemble the student's self-explanations may produce better reasoning performance. We then propose \teachback{}, demonstrating that the in-context teaching ability of LLMs can be improved by aligning the teacher with the student's self-explanations.
\end{itemize}





\begin{figure*}[t]
 \centering
 \includegraphics[scale=0.44]{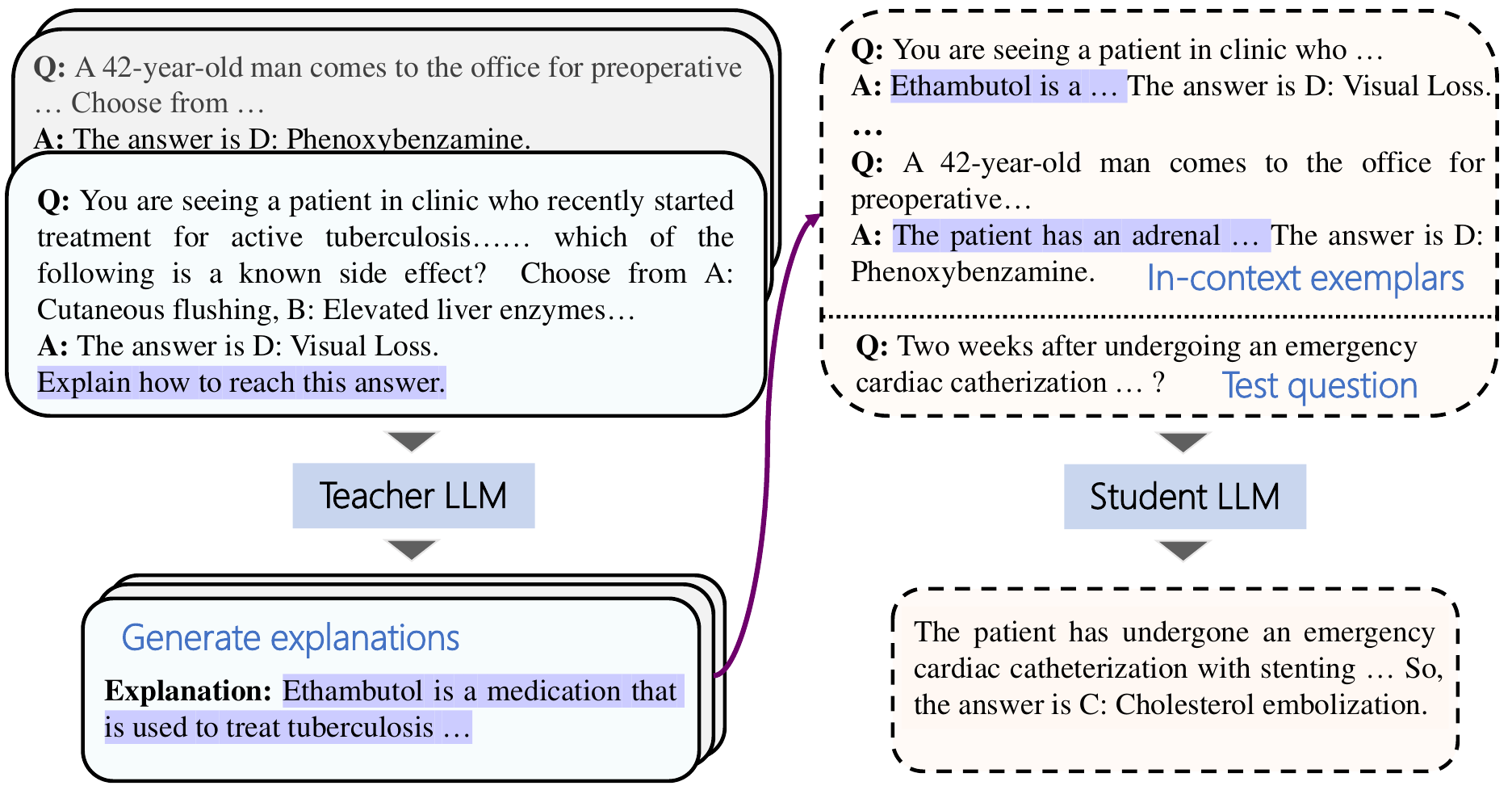}
 \caption{ The overall framework and prompting format of our approach. The teacher LLM is prompted to generate explanations on sampled training data. Those teacher's explanations are used as in-context demonstrations for the student model at test time. The student model and the teacher model can be the same. } \vspace{-2mm}
 \label{fig:pipeline}
\end{figure*}

\section{Revisiting ICL}
 We first detail some annotations and give a formal setup of in-context learning. We denote the model parameters as $\theta$, rationale as $\pi$ and assume a labeled dataset $\mathcal{D}$ with distribution $p^{\ast}$. Given a test query $\mathbf{x}$, the model will predict $\tilde{\mathbf{y}}$ by conditioning on the query and in-context exemplars.  We can then have,
\begin{equation}
    \tilde{\mathbf{y}}=\text{argmax}_{y}P(y|\mathbf{e},\mathbf{x},\theta), \label{eq:icl}
\end{equation}
where $\mathbf{e}$, is the sequence of all $K$ in-context exemplars i.e., $\mathbf{e}=e_{1},...,e_{K}$) and $e_{i}=(x_{i},\pi_{i},y_{i})$ where $(x_{i},y_{i})$ is sampled from $p^{\ast}$.

\subsection{Encoding Specificity Hypothesis} 
A key question for ICL is {how to compose in-context rationales for some task dataset} $\mathcal{D}_{(x,y)}$? 
Rationales can be rephrased differently while delivering the same logic.
To understand this question, we take a memory view of ICL by conceptualizing LLMs as memory networks~\cite{hopfield1982neural,kanerva1988sparse,kaiser2016can,ramsauer2020hopfield, krotov2016dense}. 
The feed-forwarding through hidden layers of LLM is to retrieve and generalize learned information in memory to construct the output $\mathbf{y}$ to complete the query $\mathbf{x}$ under the guidance of context $\mathbf{C}$ (i.e., in-context exemplars). 
The pretraining stage can be viewed as encoding information into the weights, i.e., memories of LLMs.  

\looseness=-1000
From a memory perspective, we draw inspiration from the encoding specificity hypothesis~\cite{tulving1973encoding}, which suggests that successful retrieval of information depends on the match between the context during encoding and the context at retrieval. To see this hypothesis, a simplified thought experiment can be considered: supposing that a specific datapoint $(x,y)$ has been seen during language modeling in pretraining and $C$ is the corresponding context prepending $(x,y)$, (i.e., a consecutive string $(C,x,y)$ is seen by LLM during training), at test time, to let the model generate $y$ with great probability, we can prompt it with $(C,x)$. 

\looseness=-1000
More generally, the encoding specificity hypothesis suggests that \textbf{in-context exemplars of reasoning at test time should match the distribution of reasoning examples seen during training}, especially the training corpus containing information similar to task data $\mathcal{D}_{(x,y)}$.  LLMs may have seen many sentences involving reasoning during pre-training and further instruction fine-tuning. It may be easier for LLMs to generalize from in-context exemplars similar to those rationales from training data (e.g., having similar reasoning logic or using similar expressions/ jargon) to answer new questions at inference.

\section{Methodology}
\label{sec:self-exp}
The general framework of our proposed methods and our prompting format is shown in Figure~\ref{fig:pipeline}. 
We first introduce \selfexp{}, where the student and teacher are the same (Section~\ref{sec:selfexp}), as a straightforward implementation of the encoding specificity hypothesis. 
We then extend this approach to employing a different teacher model (Section~\ref{sec:teachback}).

\subsection{\selfexp{}}\label{sec:selfexp}
\looseness=-1000
Motivated by the encoding specificity, we would like our in-context exemplars of reasoning to match the LLMs' training corpus containing information similar to the task data. To achieve this, we directly prompt an LLM to elicit its explanation for some question-answer pairs of task data. 
Such self-explaining is actually how humans integrate new information with their existing knowledge~\citep{chi1989self}. Similarly, the LLM is expected to utilize its existing encoded knowledge relevant to the unseen question provided, in order to generate its explanations. These self-explanations are then used as in-context exemplars of reasoning to prompt the model itself. 

\paragraph{Eliciting LLMs' Self-explanations.} Formally, we assume access to labeled training data where we have some data $(x,y)$ sampled from the distribution $p^{\text{train}}$ and assume $p^{\text{test}}\approx p^{\text{train}}$. We consider a realistic setting where human-crafted CoTs are not available.  We define an oracle CoT as 
\begin{align}
    \pi^{\ast}:= \text{argmax}_{\pi}P(y|x,\pi,\theta).
\end{align}

Self-explanation is then obtained as,
 \begin{align}
\pi^{\text{self}}=\text{argmax}_{\pi}P(\pi|x,y,\gamma,\theta), \label{eq:generate_exp}
 \end{align}
 \looseness=-1000
 where $\gamma$ is an instruction. We hope LLMs generate rationales based on a given $(x,y)$ by recalling relevant encoded knowledge so as to satisfy the encoding specificity. We find $P(y|x,\pi^{\text{self}},\theta)>>P(y|x,\pi^{\text{human}},\theta)$ (see Appendix~\ref{sec:conf}). We may arguably state that $\pi^{\text{self}}$ is a more reasonable estimation to $\pi^{\ast}$ than  $\pi^{\text{human}}$.

 \paragraph{Filtering Self-explanations.}   We filter out the elicited self-explanations based on the explanation \textit{faithfulness}~\citep{jacovi-goldberg-2020-towards}. Explanations that fail to guide the model to produce the given answer, i.e., $y \neq \text{argmax}_{\tilde{y}}P(\tilde{y}|x,\pi^{\text{self}},\theta)$, are screened.  We empirically verify the self-explanation ability of different LMs and show that those models succeed in justifying the given $(x,y)$ most of the time (see Section~\ref{sec:faithexp}).



\paragraph{ICL with Self-explanations.}  \label{sec:method-exp-icl}
\looseness=-1000
The self-explanations $\pi^{\text{self}}$ elicited by the model are then used as in-context exemplars for prompting it following Equation~\ref{eq:icl}. This can be viewed as the model teaching itself via ICL to do reasoning. Additionally, the ICL performance is very close when using respective $\pi^{\text{self}}$ elicited with either wrong or ground-truth $y$ for the input question $x$ in Equation~\ref{eq:generate_exp} (see Section~\ref{para:correct_self_exp}). 


\paragraph{Generalization through generation diversity.}\label{para:multi}  The underlying logic of $\pi^{\text{self}}$ might be very specific to its corresponding $(x,y)$ and thus lacks generalizability to other different cases from test data. Then, the output explanation $\hat{\pi}_{\text{te}}$ at test time may fail to apply to the input cases, leading to wrong answers.  To mitigate this issue, we also design a new instruction $\gamma'$ so as to prompt the model to generate solutions employing distinct logics.  Formally, we have, 
\begin{equation}
    (\pi^{\text{self}}_{1},...,\pi^{\text{self}}_{n})=\text{argmax}_{\pi}P(\pi|x,y,\gamma',\theta),
\end{equation}
where $n\in (1,N)$ and $N$ is the number of different explanations to generate. For example, if $N=5$, $\gamma'$ will be ``Explain how to reach this answer in five different ways''. Then at test time, $\pi^{\text{self}}_{i}$ for an in-context exemplar $(x_{i},y_{i})$  will be randomly sampled from the according $\{\pi^{\text{self}}_{n} | n\in (1,N)\}$ of $(x_{i},y_{i})$.

\subsection{\teachback{}}\label{sec:teachback}

Instead of ICL with self-explanations where the student model teaches itself to reason, these explanations can be provided as in-context exemplars by a different model (parameterized by $\theta_{\text{teacher}}$), i.e.,
\begin{align}
  \pi^{\text{self}}=\text{argmax}_{\pi}P(\pi|x,y,\gamma,\theta_{\text{teacher}}).  
\end{align}

However, explanations of one model may not be the most helpful reasoning demonstrations for another model, especially when the teacher's explanations are very distinct from the student's self-explanations (see results in Section~\ref{sec:ret-teaching}). 
Based on our encoding specificity hypothesis,  we propose to let the teacher model learn from the student's self-explanations (through supervised fine-tuning) before eliciting the teacher's explanations.
This method is called \teachback{}, which is similar to how healthcare providers reduce the communication gap with patients for effective health education~\citep{talevski2020teach}.  
Doctors will rephrase and clarify their explanations based on patients' explanations for better communication. 
In Section~\ref{sec:teach-back}, we empirically show the effectiveness of \teachback{} in improving teaching efficacy and enhancing student's performance.

\begin{table*}[t]
\centering
\renewcommand{\arraystretch}{1.3}
\begin{tabular}{ccccc}

\hline\toprule
\textbf{Method}\textbackslash \textbf{Dataset} & \textbf{MedCQA} & \textbf{MedQA} & \textbf{StrategyQA} & \textbf{}\\ \hline
No CoT                       &    51.7   & 52.1     &46.8& \\
Zero-shot CoT~\citep{kojima2022large}                & 51.1      &54.4      &45.6 \\
Auto-CoT~\citep{zhang2023automatic} &52.5 &55.2 & 52.7\\
Human CoT                    &    53.1   &55.6    &56.1& \\ \hdashline
\selfexp{}              & 53.2     & 57.5      &58.5 &\\
\multicolumn{1}{r}{w/ Multi-Exp}              &    \textbf{56.6}  &    \textbf{59.6}    &\textbf{59.7}&\\ 
\bottomrule
\end{tabular}
\caption{Test accuracy of different prompting methods on three datasets for knowledge reasoning.}
\label{tab:main-ret}
\end{table*}




\section{Experiments}
\subsection{Experimental Setup}
\paragraph{Datasets.}\looseness=-1
We focus on knowledge-intensive question-answering tasks that require logical reasoning and associating encoded knowledge, rather than just retrieving facts.  Such knowledge-intensive QA is common and important for the applications of LLMs~\cite{Jin2021BiomedicalQA, Tran2023BioInstructIT}. We evaluate our method in both general domains and expert domains. We employ widely-used StrategyQA~\cite{geva2021did} for commonsense reasoning. For expert domains, we use challenging MedMCQA~\cite{pal2022medmcqa} and MedQA~\cite{jin2021disease} with standard splits. These datasets consist of multiple-choice questions to diagnose clinical cases, which are used for physician qualification exams.

\vspace{-2mm}
\paragraph{Models.} We use a variety of language models. We employ the chat version of 7B model and 13B model of Llama2~\citep{touvron2023llama}, the 7B model of Mistral~\citep{jiang2023mistral}, the Phi3-128k-mini that has 3.8B parameters~\cite{abdin2024phi} and the frozen version (0613) of GPT-3.5~\footnote{https://platform.openai.com/docs/models/gpt-3-5-turbo}.

\vspace{-2mm}
\paragraph{Prompting.}   For the instruction used for eliciting models' self-explanations, an ablation study is conducted in Appendix~\ref{app:effect_cue}. For few-shot prompting at test time, we use five in-context exemplars sampled from the training data.

\vspace{-2mm}
\paragraph{Baselines.}
Apart from comparing our approach with standard few-shot prompting with human CoTs, we include three more baselines. (1) ``No CoT'': We remove rationales and use input-output pairs only for in-context exemplars; (2) ''Zero-shot CoT''~\citep{kojima2022large}: This method does not require human-crafted demonstrations as it is not few-shot prompting.
It directly elicits reasoning from LLMs for the test question by using the prompt ``Let's think step by step''. (3) ``Auto-CoT''~\citep{zhang2023automatic}: This work uses the same method as \citet{kojima2022large} to elicit rationales from LLMs.
But it further proposes a way of exemplar selection to choose elicited rationales as in-context exemplars. For fair comparison, in each trial, we use the same question-answer pairs for few-shot demonstrations for all baselines.

\begin{table}[t]
\centering
\resizebox{0.45\textwidth}{!}{
\renewcommand{\arraystretch}{1.3}
\begin{tabular}{ccccc}
\hline
 & \textbf{MedMCQA} & \textbf{MedQA} & \textbf{StrategyQA} &\\ \hline
Right                       &    56.6   & 59.6    &59.7 \\
Wrong               &    56.0   &59.4    &59.1\\ 
\bottomrule
\end{tabular}
}
\caption{Test accuracy of prompting with self-explanations that are generated provided by right answers and wrong answers. }
\label{tab:right-wrong-effect}
\end{table}

\begin{figure}[t]
    \centering
    \includegraphics[scale=0.4]{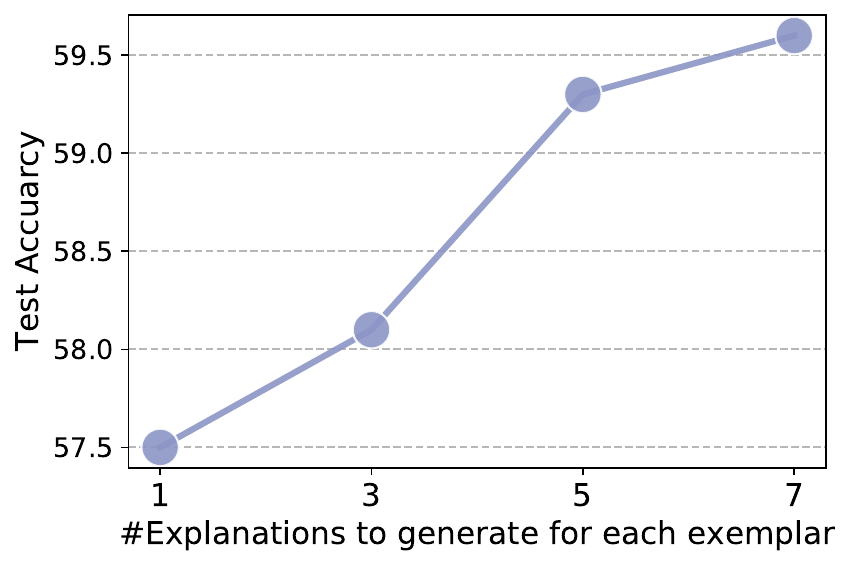}
    \caption{The test performance with respect to the number of self-explanations to generate for each exemplar.}\vspace{-5mm}
    \label{fig:num-to-generate}
\end{figure}

 \begin{table*}[t]
\centering
\begin{tabular}{lccccc}\toprule
\textbf{Teacher\textbackslash Student} & \textbf{Llama2-7B} & \textbf{Llama2-13B} & \textbf{Mistral-7B} & \textbf{Phi3-mini} & \textbf{GPT-3.5} \\ \hline
\textbf{No CoT}                        & 28.4               & 31.1                & 31.8                & 49.5               & 41.1             \\
\textbf{Human}                         & 27.3               & 31.4                & 38.2                & 53.3               & 55.6             \\ \hdashline
\textbf{Llama2-7B}                     & \textbf{30.6}      & 32.2                & 40.8                & 49.1               & 51.2             \\
\textbf{Llama2-13B}                    & 30.2               & \textbf{35.5}       & 41.1                & 55.3               & 56.9             \\
\textbf{Mistral-7B}                    & 25.1               & 34.7                & \textbf{44.2}       & 54.4               & 53.5             \\
\textbf{Phi3-mini}                     & 18.7               & 35.1                & 40.7                & 57.1               & 57.1             \\
\textbf{GPT-3.5}                       & 18.1               & 34.4                & 43.1                & \textbf{57.7}      & \textbf{57.5} \\
\bottomrule
\end{tabular}
\caption{Results of teaching student LLMs with teachers' self-explanations through in-context learning. The best test accuracy is highlighted in bold.}\vspace{-5mm}
\label{tab:in-context-teach}
\end{table*}

\begin{figure*}[t]
  \centering
  \begin{subfigure}{0.45\linewidth}
    \includegraphics[width=\linewidth]{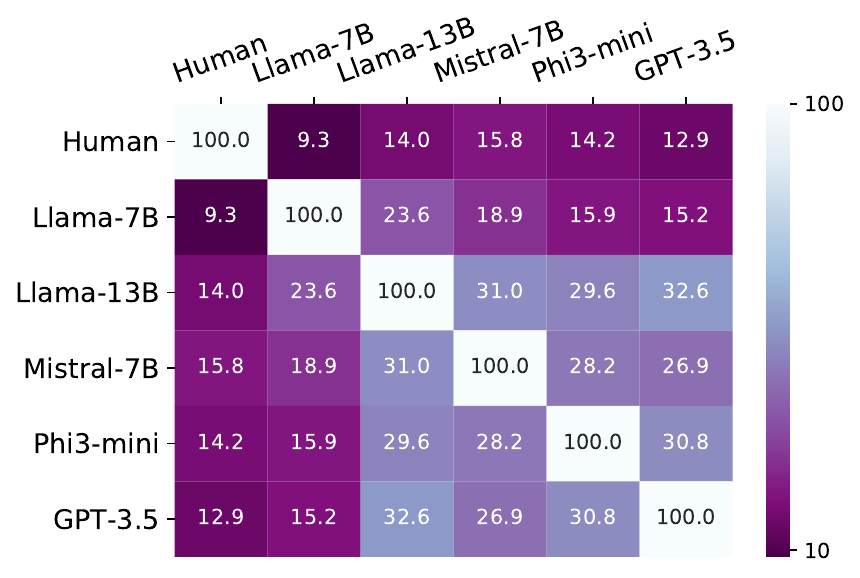}
    \caption{}
    \label{subfig:rouge}
  \end{subfigure}%
  \hfill
  \begin{subfigure}{0.4\linewidth}
    \includegraphics[width=\linewidth]{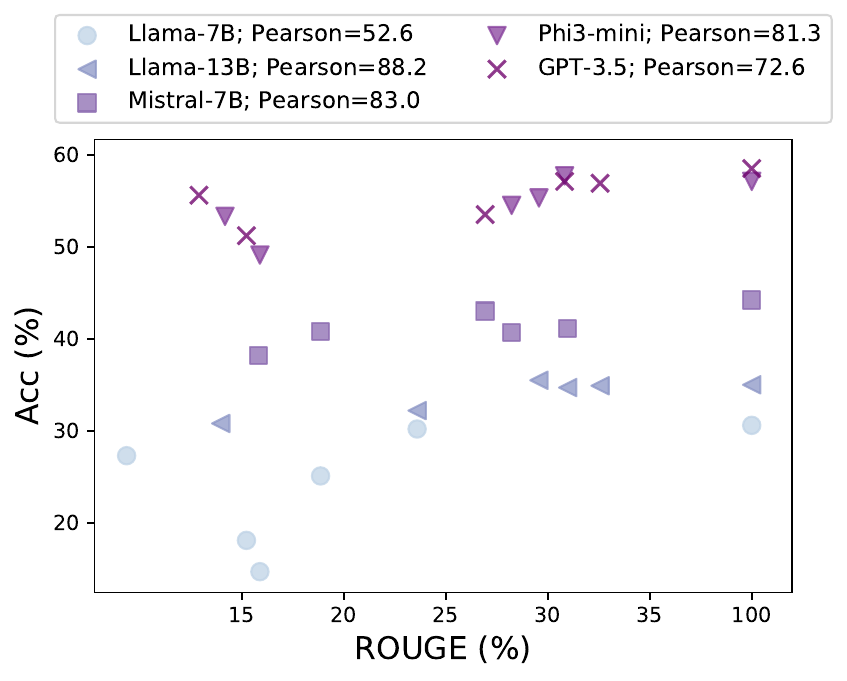}
    \caption{}
    \label{subfig:rouge-acc}
  \end{subfigure}%
\caption{(a) ROUGE scores between self-explanations of teacher and student. For ``Human'' teachers, human-crafted CoTs are used for computation. (b) Strong linear correlation is observed between ROUGE scores of self-explanations of teacher and student and the student's test accuracy.}\vspace{-5mm}
\end{figure*}

\subsection{Few-shot Prompting with Self-explanations}\label{sec:few-prompt-imp}
In this section, we evaluate the test performance of \selfexp{}, i.e., when the model's self-explanations are used as in-context exemplars of reasoning for few-shot prompting. We conduct experiments with GPT-3.5-turbo on reasoning tasks in both general domains and expert domains.

\paragraph{Prompting with self-explanations is better than using human-crafted CoTs.}\looseness=-1000  Our results are shown in Table~\ref{tab:main-ret}. \selfexp{} can impressively outperform using CoTs crafted by human professionals by around 2\% in both challenging MedQA and general domain, while reaching similar performance to Human CoT for MedCQA.  Our approach also outperforms Auto-CoT~\citep{zhang2023automatic} and vanilla zero-shot CoT~\citep{kojima2022large}, both of which cannot effectively surpass Human CoT for knowledge reasoning. The superior performance of \selfexp{} may support our encoding specificity hypothesis. Overall, considering the difficulty and expense of crafting CoTs by humans, \selfexp{} can thus be very useful in expert domains. Example self-explanations and human-crafted CoTs can be found in Appendix~\ref{app:examp}.



\paragraph{Generation diversity is helpful.}  
Apart from naive \selfexp{}, we generate five different self-explanations for each in-context exemplar and randomly select one for ICL at test time (see details in Section~\ref{para:multi}). As shown in results of ``w/ Multi-Exp'' in Table~\ref{tab:main-ret}, this approach further boosts the performance of \selfexp{} to significantly surpass Human CoT by around 4\% in all datasets.  To better understand the effects of this component, we experiment with generating different numbers of self-explanations for one exemplar input. Results are shown in Figure~\ref{fig:num-to-generate}.  We find generating different self-explanations for an in-context exemplar can generally improve the test performance, while such improvement experiences diminishing returns with further increased numbers of generations.

\paragraph{Does the correctness of self-explanations matter?}\label{para:correct_self_exp}\looseness=-1000
A natural question raised in \selfexp{} is what if the self-generated explanations are wrong since the generation process is not supervised by humans. We look into this question by providing the LLM with random wrong answers to generate misleading explanations of the input question (i.e., we use $(x,y_{\text{wrong}})$ in Equation~\ref{eq:generate_exp}).  Those self-explanations with wrong answers i.e., $(x,\pi^{\text{self}}_{\text{wrong}},y_{\text{wrong}}$) are then used for prompting as in-context exemplars.  The results are shown in Table~\ref{tab:right-wrong-effect}. 
We find that the performance of prompting with self-explanation seems insensitive to its correctness. 
This result suggests that a correctly labeled dataset may not be necessary for \selfexp{} prompting. 
Similar results on text classification are observed that label space is more important for ICL than label correctness~\citep{min-etal-2022-rethinking}.
We similarly speculate that what carries more weight is how self-explanations are phrased, as they should match the context seen during encoding relevant information according to our encoding specificity hypothesis. 
We look deeper into this hypothesis in Section~\ref{sec:ret-teaching}.  


\subsection{In-context Teaching via Explanations}\label{sec:ret-teaching}

We have demonstrated LLMs can teach themselves with \selfexp{} for better knowledge reasoning. 
We further extend this to study whether self-explanations of one model can be used as in-context exemplars to teach another model through ICL.  
Teaching through supervised learning on teacher's generated data has been widely investigated~\citep{distill24,ho-etal-2023-large,hsieh-etal-2023-distilling}, which can be framed as knowledge distillation.  
However, machine supervision through ICL has not yet been well studied. 
In this section, we have a teacher LLM generate self-explanations that are then used as in-context exemplars to teach a student LLM for reasoning unseen test cases. 
\citet{saha2023can} have explored a similar research question, while they insert teacher's explanations into student's generation for test examples during inference. 
This may not be fully considered as \textit{teaching} as the taught model receives assistance with test examples, and its generalization ability is thus not evaluated. 

\vspace{-2mm}
\paragraph{The student is often its own best teacher.}\looseness=-1000
Results are shown in Table~\ref{tab:in-context-teach}.
When doing few-shot prompting with the students' own self-explanations as in-context exemplars, the students can generally reach the best performance, which is aligned with results in Section~\ref{sec:few-prompt-imp}. 
This also supports the encoding specificity hypothesis. 
Noticeably, larger or stronger models may not necessarily be better in-context teachers. 
For example, for Llama2 and Mistral, prompting them with GPT-3.5's explanations gives worse results than using those models' own self-explanations. 
The performance for Llama2-7B with GPT-3.5 as the in-context teacher is even worse than not using any demonstrations of reasoning (i.e., ``No CoT'').

\paragraph{Better teachers tend to produce rationales that more resemble student's self-explanations.} The performance or scale of the teacher model is shown not indicative of its in-context teaching ability.
Instead, we observe a strong correlation between students' performance and the similarity between teachers' demonstrations and students' self-explanations. 
Specifically, ROUGE-L~\citep{lin-2004-rouge} is used to measure the closeness between rationales of teachers and students.
We observe that explanations generated by different LLMs tend to be more similar to each other than to human-crafted rationales, as shown in Figure~\ref{subfig:rouge}. 
Meanwhile, LLM teachers tend to yield better student's performance than human teachers shown in Table~\ref{tab:in-context-teach}. 
We further compute Pearson correlation coefficient between the student LLM's test accuracy and the ROUGE score between its self-explanations and the teacher LLMs' explanations/ human-crafted rationales. 
As shown in Figure~\ref{subfig:rouge-acc}, evident linear correlations between accuracy and ROUGE score are observed, especially for the models stronger than Llama2-7B. 
This may further support our encoding specificity hypothesis: in-context exemplars for the student should match its training data because rationales similar to the student's self-explanations are more likely to align with the student's training data. 

\begin{figure}[t]
    \centering
    \includegraphics[scale=0.5]{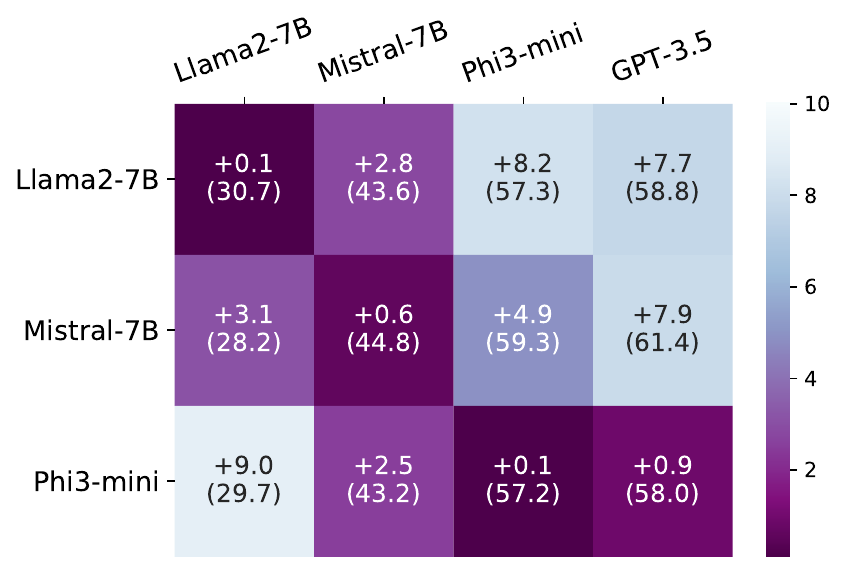}
    \caption{Students' accuracy improvement after applying \teachback{}. Values in brackets stand for students' respective test accuracy w/ \teachback{}.  The x-axis represents students who do reasoning. The y-axis is teacher models that provide in-context demonstrations for students. }\vspace{-5mm}
    \label{fig:sft-imp-acc}
\end{figure}

\subsection{Learning from Students for Better Teaching}\label{sec:teach-back}
Given the strong correlation between students' test accuracy and closeness between teacher and student explanations, we are motivated to further look into the underlying causality, i.e., \textbf{whether in-context teaching can be improved by letting the teacher learn the student's self-explanations}. 
In this section, the teacher model will first be fine-tuned on the student’s self-explanations to generate its new self-explanations that are then used as in-context exemplars for the student (i.e., \teachback{} introduced in Section~\ref{sec:teachback}).

Each student model generates self-explanations for 500 held-out training examples (will not be used for in-context demonstrations) for fine-tuning teacher models. 
To accommodate our available computing resources, we only fine-tune the teacher models whose size is smaller or equal to 7B with LoRA~\citep{hu2021lora}. Detailed implementations for fine-tuning are shown in Appendix~\ref{app:detail-exp-teachback}. Example generations before and after \teachback{} are shown in Appendix~\ref{app:example-diff-llm}.

\paragraph{\teachback{} improves in-context teaching.}
\looseness=-1000
As shown in Figure~\ref{fig:sft-imp-acc}, when the teacher model is different from the student model, \teachback{} can greatly enhance the teaching performance of the teacher LLM, as evidenced by the improvement of test accuracy among students.  
Noticeably, fine-tuned teachers using \teachback{} can enable students to achieve significantly higher accuracy than the former best teachers in Table~\ref{tab:in-context-teach}.  
For example, a fine-tuned Mistral-7B can guide Phi3-mini to achieve 59.3\% accuracy.
This is 4.9\% higher than the accuracy achieved with an unfine-tuned Mistral-7B teacher and 1.6\% higher than the best unfine-tuned teacher (i.e., GPT-3.5, see the column for ``Phi3-mini'' in Table~\ref{tab:in-context-teach}).  
Interestingly, \teachback{} enables the smaller Mistral-7B to teach the much larger GPT-3.5 in context, surpassing human teachers by around 5\% and \selfexp{} by around 4\% as visualized in Figure~\ref{fig:cmp-huma-self-teach-back} of Appendix.  
Our results showcase the promising use of \teachback{} in leveraging a small tunable model to improve the few-shot prompting performance of a much larger LM without human supervision (i.e., human-crafted demonstrations). 
\begin{figure*}[t]
    \centering
    \includegraphics[scale=0.46]{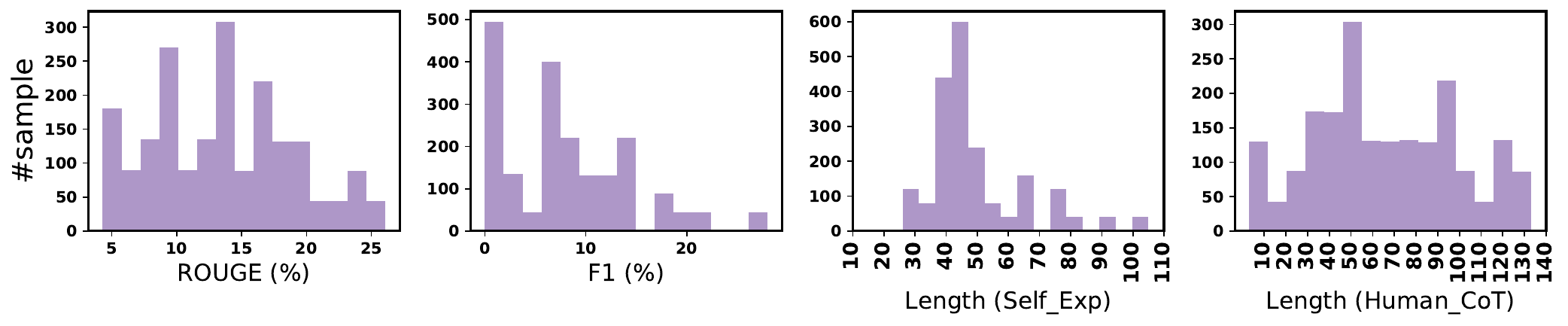}
    \caption{Similarity between human-crafted CoTs and self-explanations in terms of ROUGE score, terminology covered, and length. }\vspace{-5mm}
    \label{fig:cmp-cot}
\end{figure*}

\section{Further Analysis of LLMs' Explanations}

\subsection{Faithfulness of Self-explanations}\label{sec:faithexp}\looseness=-1000
To elicit self-explanations $\pi^{\text{self}}$ from an LLM, we prompt it to explain a given pair of question and answer $(x,y)$ as shown in Equation~\ref{eq:generate_exp}, and then $(x,\pi^{\text{self}},y)$ will be used as one in-context demonstration.  
In this section, we evaluate how many of those raw self-explanations (before filtering) actually support the model to predict the given answer~\citep{hase-etal-2020-leakage}.
We append the elicited explanations $\pi^{\text{self}}$ back to the given question $x$ as the prompt fed to the model. We then examine whether the model will correctly output the given answer $y$. We empirically find that diverse models can produce faithful explanations most of the time. For example, Mistral-7B reaches 94.2 \% rate of faithful explanations, GPT 3.5 reaches 98.3\% and Llama2-13B reaches 93.9 \%.  Our results may confirm the LLM's ability to explain given questions and answers.


\subsection{How Similar are Self-explanations to Human-crafted CoTs?}
The common standard to measure the quality of machine-generated samples is how similar they are to human-crafted ones~\citep{lu2022learn,wang2022self}. 
The more similar, the better the quality is assumed to be. However, \citet{hase-etal-2020-leakage} have pointed out that evaluation based on plausibility by matching human explanations is not sufficient. Our results also challenge this evaluation criterion. 
We show that LLMs' self-explanations are very different from human-crafted CoTs in terms of ROUGE-L score, terminology used, and length.  
However, few-shot prompting with LLMs' self-explanations demonstrates superior performance to using human-crafted CoTs.

We use MedCQA as our testbed, which provides high-quality human-crafted explanations.
For terminology comparison, we extract terms in both kinds of CoTs through scispaCy~\footnote{https://allenai.github.io/scispacy/} and calculate F1 score between the two terms lists. 
Results are shown in Fig.~\ref{fig:cmp-cot}.  We find in terms of content (measured by ROUGE-L and term coverage), self-explanation differs from human-crafted CoTs greatly, given the average similarity is around 15\%. 
The length distribution of self-explanation is more centric, while the human-crafted CoTs have more varied lengths.

\vspace{-2mm}

\section{Related Work}\label{sec:relatework}

\paragraph{In-context learning.}
In-context learning is the ability of language models to induce answers from given demonstrations without weights updating in supervised tuning. 
In-context exemplars are the key to ICL which have dominating influence on the generation.
Quite a few works have been proposed to optimize the selection of exemplars~\citep{lu2023makes,rubin-etal-2022-learning,DBLP:conf/iclr/FuPSCK23}. 
On the other hand, in the cases of no access to task labels, \citet{lyu-etal-2023-z} proposed a zero-shot ICL that employs pseudo exemplars with random labels for classification tasks. 


\paragraph{Chain-of-Thought prompting without human-crafted exemplars.}\looseness=-1000  Prompting with reasoning exemplars triggers LLMs to generate similar intermediate steps of thinking through ICL, known as Chain-of-Thought (CoT)~\citep{DBLP:conf/nips/Wei0SBIXCLZ22}. 
\citet{kojima2022large} propose zero-shot CoT prompting to elicit LLMs' reasoning without human-crafted exemplars. 
This method is then leveraged to prompt LLMs to generate CoT exemplars by themselves for ICL~\citep{zhang2023automatic,wan2023better,chen2023self}.
Different from our work, which focuses on eliciting rationales from LLMs, \citet{zhang2023automatic,wan2023better,shum-etal-2023-automatic} concentrate on selecting rationales generated according to \citet{kojima2022large}. 
And \citet{chen2023self} further incorporate pseudo task generation alongside self-generated CoTs.
Additionally, \citet{yasunaga2024large} propose analogical prompting to solve emerging new tasks without human-crafted demonstrations. 
Importantly, these works mainly focus on prompting engineering for very large, closed-source LMs (e.g., GPT-4). 
None of them formally investigate the fundamental in-context teaching among different LMs. 
Instead, our work proposes encoding specificity hypothesis to understand in-context teaching for LLMs, which is evidenced by experiments across different models.



\paragraph{Teaching via explanations.} \looseness=-1000
Many past works have explored teaching student LLMs with teacher model's explanations through supervised fine-tuning~\citep{ho-etal-2023-large,hsieh-etal-2023-distilling,fu2023specializing}. 
Few have investigated in-context teaching. \citet{lampinen-etal-2022-language} demonstrate that LLMs can learn from human-crafted explanations in context. Instead of leveraging in-context exemplars, \citet{saha2023can,lee-etal-2024-small} directly feed the teacher's explanations of test examples to the student model during inference. In this case, the taught model receives direct assistance with test examples, and thus, its generalization ability from the teacher is not evaluated. 


\section{Conclusion}
In this work, we investigate in-context teaching, where a teacher provides in-context
example rationales to teach a student to reason over unseen questions. We introduce the encoding specificity hypothesis that in-context exemplars at test time should match the student model’s related training examples.  Motivated by our hypothesis, we propose \selfexp{} to let an LLM teach itself with its self-explanations through in-context learning, which outperforms human-crafted chain-of-thoughts and other baselines in different reasoning tasks. We reveal that for in-context teaching, rationales by distinct teacher LLMs or human teachers that more resemble the student LLM's self-explanations are better demonstrations, which further supports the encoding specificity hypothesis. 
We then propose \teachback{} to align the teacher LLM with the student, which can enhance the in-context teaching performance.

\section{Limitations}
We propose \selfexp{} and \teachback{} that verify our encoding specificity hypothesis for few-shot prompting. They also demonstrate impressive performance on diverse models for knowledge reasoning without human guidance.
The student model's performance with \selfexp{} is consistently among the best. However, the student performance in \teachback{} does not necessarily surpass standard prompting with human CoTs, depending on the teacher model. 
For example, Mistral-7B with \teachback{} enables different student models to reach optimal test performance, while teachers like Llama2-7B are less effective. Therefore, we suggest using \selfexp{} as a starting point in real applications. In the future, we will further investigate the influence of the teacher model in \teachback{} on student performance and how fine-tuning affects the teacher model's self-explanations. Overall, in this work, the main contribution of our proposed \teachback{} is that it can greatly improve the ability of one LLM to teach a different student model.

In addition, our work is limited to only one teacher. Future work could explore many teachers, including mixture of experts. Moreover, there are various emerging advanced prompting methods for different kinds of reasoning tasks, e.g., tree-of-thoughts~\citep{yao2024tree} or multi-round prompting~\citep{DBLP:conf/iclr/KhotTFF0CS23,zhou2022least}. In this work, we do not consider these more advanced designs of prompting, but focus on commonly used CoT prompting to eliminate the need of human-crafted CoTs.
However, our approaches can be adapted to these methods e.g., by modifying the instructions used to elicit LLMs' rationales. The majority of these methods still require human-crafted demonstrations.
We will further investigate whether LLMs can implement these advanced prompting methods without human-crafted exemplars under our framework in the future.






\bibliography{ref}
\newpage
\appendix

\begin{figure}[t]
    \centering
    \includegraphics[scale=0.5]{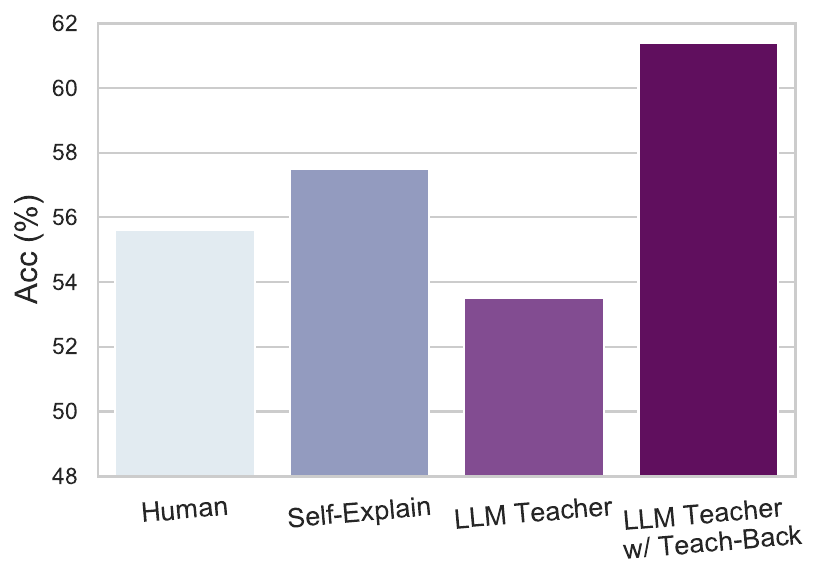}\vspace{-3mm}
    \looseness=-1
    \caption{ Few-shot prompting performance of GPT-3.5 on knowledge reasoning with different kinds of in-context exemplars of reasoning. ``Human'' is using chain-of-thought examples crafted by humans.  ``Self-Explain'' is prompting GPT-3.5 with its self-explanations elicited. ''LLM Teacher'' is using rationales generated by a separate model (Mistral-7B) as teacher, while for ``w/ Teach-Back'', the teacher model has been first fine-tuned with GPT-3.5's self-explanations.  }\vspace{-4mm}
    \label{fig:cmp-huma-self-teach-back}
\end{figure}

\begin{table*}[t]
\centering
\begin{tabular}{rll}
\toprule
  & \multicolumn{1}{c}{\textbf{Medical Domain}} & \multicolumn{1}{c}{\textbf{General Domain}} \\\hline 
1 & \multicolumn{2}{c}{Explain how to reach this answer.}\\
2 & \multicolumn{2}{c}{Let's think step by step.}\\
3 & Let's think step by step like a medical expert. &  Let's think step by step like an expert.\\
4 & Let’s use step by step inductive reasoning,  &  Let’s use step by step inductive reasoning. \\
&given the medical nature of the question.&\\
\bottomrule
\end{tabular}
\caption{Different cues to elicit self-explanations.}
\label{tab:cues}
\end{table*}

\begin{table*}[t]
\centering
\begin{tabular}{cccccc}\toprule
\textbf{Dataset} & \textbf{Cue \#1} & \textbf{Cue \#2} & \textbf{Cue \#3} & \textbf{Cue \#4} & \textbf{Human}\\ \hline
  MedMCQA      &56.6          &54.6            & 54.3    &  54.2   &    53.1\\
  MedQA      & 59.6         & 59.4           & 58.1    & 58.2   &    55.6\\
  StrategyQA &59.7 & 57.7 &57.2 &57.3 &56.1\\
\bottomrule
\end{tabular}
\caption{Test results of ICL with self-explanations elicited by different cues.}
\label{tab:cues_result}
\end{table*}


\begin{table}[htp]
\centering
\begin{tabular}{l}
\toprule
\textbf{Input}:\{input string of training example $i$\}\\
\textbf{Output}:\{output result of training example $i$\}\\
Explain how to reach this answer.\\
\{explanation for training example $i$\}\\
\bottomrule
\end{tabular}
\caption{The format of training data for fine-tuning teacher LLMs on students' self-explanations.}
\label{tab:sft-format}
\end{table}

\section{Effects of Instructions in Eliciting Self-explanations}
\label{app:effect_cue}
In this section, we examine the performance of ICL with self-explanations prompted by different cues in our framework. We mainly follow cues in \citet{lievin2022can} as shown in Table~\ref{tab:cues}. The first one is by default used in our framework.  Since \citet{lievin2022can} focuses on medical domains, for general domains, we modify its cues by removing information specific to medical domains. We then generate self-explanations and perform ICL with them.  The final test results are shown in Table~\ref{tab:cues_result}.  We find no matter what cues are employed, ICL with self-generations elicited can all outperform using human-crafted CoTs, which demonstrates the robustness of our proposed \selfexp{} on the choice of cues.

\section{Implementation for \teachback{} with Fine-tuning}\label{app:detail-exp-teachback}
We reformat the training data with students' self-explanations following the template in Table~\ref{tab:sft-format}. We set the learning rate as $1 \times 10^{-5}$ and fine-tune the teacher model with five epochs. We use the default setting for LoRA.

\begin{figure*}[t]
  \centering
  \begin{subfigure}{0.4\linewidth}
    \includegraphics[width=\linewidth]{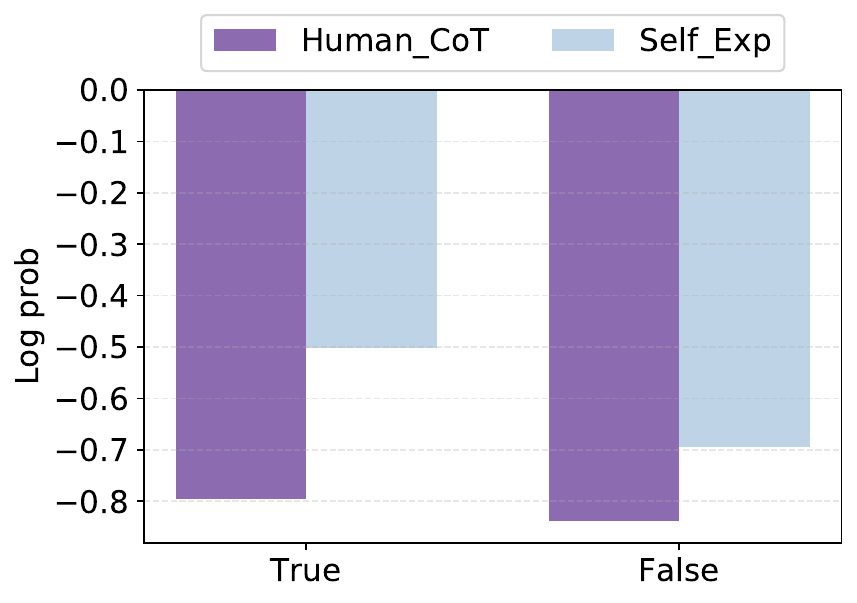}
    \caption{}
    \label{fig:cali1}
  \end{subfigure}%
  ~
  \begin{subfigure}{0.4\linewidth}
    \includegraphics[width=\linewidth]{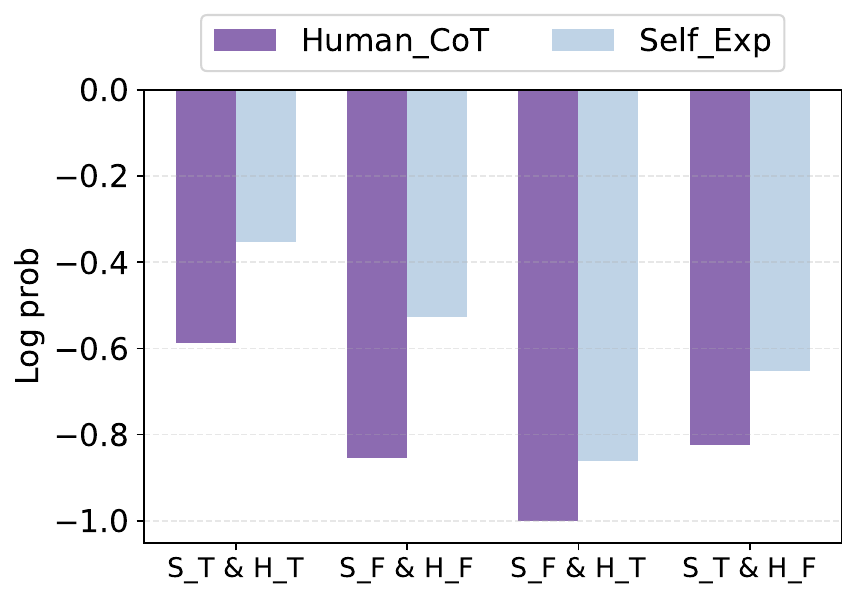}
    \caption{}
    \label{fig:cali2}
  \end{subfigure}%
  \caption{Comparison of model confidence in different cases by using human-crafted CoT and self-explanation.  Prompting with self-explanations makes the model more calibrated and more confident as well. \textbf{(a)} Comparison between confidence of true answers and false ones.  \textbf{(b)} Further comparison of confidence in more specific cases. ``S'' denotes ``Self\_exp'', ``H'' denotes ``Human\_cot'', ``T'' denotes ``True'' and ``F'' denotes ``False'', e.g., ``S\_T \& H\_F'' means self explanation gives true answer, while human-crafted CoT leads to false answer. }\vspace{-5mm}
  \label{fig:conf_cali}
\end{figure*}

\section{Analysis on Model Confidence}\label{sec:conf}

We find models are more confident with their predictions when using \selfexp{}.  For exemplars selected for generating self-explanation, we use Text-Davinci-003 to compute the average  $P(y|x,\pi^{\text{self}},\theta)$ which reaches 99.96\%. In comparison, for human-crafted explanation of the same exemplars, average $P(y|x,\pi^{\text{human}},\theta)$ is lower, reaching 89.05\%. This implies that for a given $(x, y)$, the self-explanation $\pi^{\text{self}}$ provides a more relevant context for eliciting $y$ than the human explanation $\pi^{\text{human}}$.

In addition, for inference with self-explanation as demonstrations in ICL, log probabilities are computed for correct and wrong model outputs prompted with self-explanation and human-crafted one. Results are shown in Figure~\ref{fig:conf_cali}.  We can observe that models' output log probabilities with self-explanation are much higher than with human-crafted explanation, indicating greater model's confidence in its output. This suggests self-explanation can be more acceptable and effective for LLMs to elicit reasoning.  \selfexp{} may also improve model's calibration.  Calibration requires the model's output confidence should indicate the correctness of answers (e.g., wrong answers have lower confidence). Well-calibrated output confidence can assist human users to determine whether to trust model's outputs or to look for alternatives in high-stakes applications.  Figure~\ref{fig:cali1} showcases the model is generally more calibrated when using self-explanation.  Further results in Fig.~\ref{fig:cali2} show that output confidence by using human CoT is not indicative especially when the question is debatable (i.e., using self-explanation and human CoT generate different answers).

\clearpage
\onecolumn
\section{Example self-explanations}
\label{app:examp}
\subsection{MedQA}
\textcolor{blue}{
Question: A 22-year-old male marathon runner presents to the office with the complaint of right-sided rib pain when he runs long distances. Physical examination reveals normal heart and lung findings and an exhalation dysfunction at ribs 4-5 on the right. Which of the following muscles or muscle groups will be most useful in correcting this dysfunction utilizing a direct method? (A) anterior scalene (B) latissimus dorsi (C) pectoralis minor (D) quadratus lumborum
\newline\newline
Self-explanation: The patient is a pregnant woman, and nitrofurantoin is considered safe for use during pregnancy. Other antibiotics, such as ciprofloxacin and doxycycline, are contraindicated in pregnancy due to potential harm to the fetus.
\newline\newline
Human-crafted explanation: We refer to Wikipedia articles on medicine for help. Among the options, only pectoralis minor muscle origins from the outer surfaces of the 3rd to 5th ribs.
\newline\newline\newline\newline
Question: A 44-year-old man comes to the office because of a 3-day history of sore throat, nonproductive cough, runny nose, and frontal headache. He says the headache is worse in the morning and ibuprofen does provide some relief. He has not had shortness of breath. Medical history is unremarkable. He takes no medications other than the ibuprofen for pain. Vital signs are temperature 37.4 C (99.4 F), pulse 88/min, respirations 18/min, and blood pressure 120/84 mm Hg. Examination of the nares shows erythematous mucous membranes. Examination of the throat shows erythema and follicular lymphoid hyperplasia on the posterior oropharynx. There is no palpable cervical adenopathy. Lungs are clear to auscultation. Which of the following is the most likely cause of this patient\u2019s symptoms? (A) Allergic rhinitis (B) Epstein-Barr virus (C) Mycoplasma pneumonia (D) Rhinovirus
\newline\newline
Self-explanation: The presentation of a fussy infant who is regurgitating all of his feeds and has yellow vomit is consistent with pyloric stenosis, which is characterized by abnormal hypertrophy of the pylorus. This condition causes narrowing of the passage between the stomach and the small intestine, leading to difficulty in food passage and subsequent vomiting.
\newline\newline
Human-crafted explanation: We refer to Wikipedia articles on medicine for help. The symptoms, especially the headache, suggest that the most likely cause is Rhinovirus. Epstein-Barr virus will cause swollen lymph nodes but there is no palpable cervical adenopathy. Lungs are clear to auscultation suggests it\u2019s not Mycoplasma pneumonia. 
}

\subsection{MedCQA}
\textcolor{blue}{
Question: Which of the following statements about Total Infra Venous Anaesthesia (TIVA) is true? (A) Causes More Renal Toxicity (B) Reduces Cerebral Metabolic Rate (C) Risk of Malignant Hyperthermia is high (D) Inhibits Hypoxic Pulmonary Vasconstriction
\newline\newline
Self-explanation: By understanding the mechanism of action: Total Intravenous Anaesthesia (TIVA) involves the administration of intravenous drugs to induce and maintain anesthesia. One of the main drugs used in TIVA is propofol, which is known to reduce cerebral metabolic rate. 
\newline\newline
Human-crafted explanation: Total intravenous anaesthesia refers to a technique in general anaesthesia using a compination of agents given solely by IV route and in the absence of all inhalation agents. TIVA (Total intravenous anaesthesia) is produced by IV propoful. Propofol decreases the cerebral metabolic rate. Propofol does not impair renal function, does not trigger malignant hyperthermia and does not inhibit hypoxic pulmonary vasoconstriction.
\newline
\newline\newline\newline
Question:  Shock causes: Choose the answer from A to D. A: Stagnant hypoxia. B: Anemic hypoxia. C: Hypoxic hypoxia. D: Histotoxic hypoxia.
\newline\newline
Self-explanation: Shock is characterized by a decrease in blood pressure and impaired circulation. This can result in reduced blood flow to the tissues, leading to stagnant hypoxia.
\newline\newline
Human-crafted explanation: Ans: A (Stagnant hypoxia)Ref: Pat GK, Hypoxia and Oxygen Therapy, In: Textbook of Medical Physiology, 2nd ed, Abuja Publishing House 2011: 94: 754-55
}

\subsection{StrategyQA}
\textcolor{blue}{
Question: Would ramen be bad for someone with heart failure?
\newline\newline
Self-explanation: Ramen can be bad for someone with heart failure due to its high sodium content. Sodium can cause fluid retention and increase blood pressure, which can put additional strain on the heart.
\newline\newline
Human-crafted explanation: People with heart failure have to limit their sodium intake. Ramen is notorious for having incredibly high sodium levels.
\newline
\newline\newline\newline
Questions: Is Cape Town south of the Equator?
\newline\newline
Self-explanation: By looking at a map: You can visually locate Cape Town on a map and observe that it is situated below the Equator, in the southern hemisphere.
\newline\newline
Human-crafted explanation:  Cape Town is an important city in South Africa. South Africa is located entirely south of the Equator.
}

\section{Examples of Different LLMs' Self-explanations}\label{app:example-diff-llm}
\textcolor{blue}{\textbf{Llama2-7B:} The patient's symptoms of decreased appetite, difficulty falling asleep, sadness, and compulsive behaviors (rechecking the door lock and counting toilet paper) are consistent with a diagnosis of obsessive-compulsive disorder (OCD). OCD is characterized by recurrent, intrusive thoughts (obsessions) and repetitive behaviors or mental acts (compulsions) that are time-consuming and cause significant distress or impairment.
\newline
The neurotransmitter serotonin is involved in the pathophysiology of OCD. Selective serotonin reuptake inhibitors (SSRIs), which increase the levels of serotonin in the synaptic cleft, are the most commonly used medications to treat OCD. }
\newline\newline

\textcolor{blue}{\textbf{Mistral-7B:} The patient's symptoms are consistent with major depressive disorder and obsessive-compulsive disorder (OCD). OCD is characterized by recurrent and intrusive thoughts (obsessions) and repetitive behaviors or mental acts (compulsions) aimed at reducing anxiety or distress caused by the obsessions. Serotonin is a neurotransmitter that plays a crucial role in regulating mood and anxiety, as well as in the modulation of obsessive-compulsive symptoms. Therefore, pharmacotherapy for this patient should be targeted towards increasing serotonin levels in the brain. Dopamine is associated with motivation and reward, glutamate with learning and memory, and norepinephrine with the fight-or-flight response.}

\subsection{Examples of Self-explanations after \teachback{}}\label{app:example-teachback}
\textcolor{blue}{
\textbf{Mistral-7B fine-tuned on Llama2-7B:} The patient's symptoms are consistent with major depressive disorder and obsessive-compulsive disorder (OCD). Both conditions are associated with alterations in neurotransmission, particularly serotonin. Serotonin plays a role in mood regulation and impulse control, which are affected in major depressive disorder and OCD, respectively. Additionally, selective serotonin reuptake inhibitors (SSRIs) are commonly used to treat both conditions. }
\newline\newline

\textcolor{blue}{\textbf{Llama2-7B fine-tuned on Mistral-7B:} The patient's symptoms are consistent with a diagnosis of obsessive-compulsive disorder (OCD). OCD is a neuropsychiatric disorder characterized by recurrent, intrusive thoughts (obsessions) and repetitive behaviors or mental acts (compulsions). The most commonly affected neurotransmitter in OCD is serotonin, which is involved in regulating mood, impulse control, and cognitive function. Therefore, pharmacotherapy targeting serotonin should be considered as a first-line treatment for the patient's symptoms.\newline The other options are not as relevant to the patient's symptoms:\newline A) Dopamine: While dopamine is involved in reward processing and motivation, it is not typically implicated in OCD.\newline B) Glutamate: Glutamate is involved in excitatory neurotransmission and is not typically associated with OCD.\newline C) Norepinephrine: While norepinephrine is involved in arousal and attention, it is not typically implicated in OCD.\newline In summary, the patient's symptoms are best explained by a diagnosis of obsessive-compulsive disorder, and pharmacotherapy targeting serotonin is the most appropriate treatment option.
}

\end{document}